# The Cafe Wall Illusion: Local and Global Perception from multiple scale to multiscale

Nasim Nematzadeh[1,2], David M.W. Powers[1]

[1]College of Science and Engineering (CSE), Flinders University, Adelaide, Australia
[2]Faculty of Mechatronics, Dept of Science and Engineering, Karaj Branch, Islamic Azad University (KIAU), Karaj-Alborz, Iran
nasim.nematzadeh, david.powers @flinders.edu.au

## ABSTRACT

Geometrical illusions are a subclass of optical illusions in which the geometrical characteristics of patterns such as orientations and angles are distorted and misperceived as the result of low- to high-level retinal/cortical processing. Modelling the detection of tilt in these illusions and their strengths as they are perceived is a challenging task computationally and leads to development of techniques that match with human performance. In this study, we present a predictive and quantitative approach for modeling foveal and peripheral vision in the induced tilt in Café Wall illusion in which parallel mortar lines between shifted rows of black and white tiles appear to converge and diverge. A bioderived filtering model for the responses of retinal/cortical simple cells to the stimulus using Difference of Gaussians is utilized with an analytic processing pipeline introduced in our previous studies to quantify the angle of tilt in the model. Here we have considered visual characteristics of *foveal and peripheral* vision in the perceived tilt in the pattern to predict different degrees of tilt in different areas of the fovea and periphery as the eye saccades to different parts of the image. The tilt analysis results from several sampling sizes and aspect ratios, modelling variant foveal views are used from our previous investigations on the local tilt, and we specifically investigate in this work, different configurations of the whole pattern modelling variant Gestalt views across multiple scales in order to provide confidence intervals around the predicted tilts. The foveal sample sets are verified and quantified using two different sampling methods. We present here a precise and quantified comparison contrasting local tilt detection in the foveal sets with a global average across all of the Café Wall configurations tested in this work.

## 1 INTRODUCTION

Visual processing starts within the retina from the photoreceptors passing the visual signal through bipolar cells to the Retinal Ganglion Cells (RGCs) whose axons carry the encoded signal to the cortex for further processing. The intervening layers incorporate several types of cell with large dendritic arbors, divided into horizontal cells that control for different illumination conditions and feedback to the receptor and bipolar cells, and amacrine cells that feed into the centre-surround organization of the Retinal Ganglion Cells. High-resolution receptors in the foveal area have a direct 1:1 pathways from photoreceptors, via bipolar cells to ganglion cells [1].

It is commonly believed that the center-surround organization in RGCs and their responses are the results of the lateral inhibitory effect in the outer and the inner retina [2] in which the activated cells inhibit the activations of nearby cells. At the first synaptic level, the lateral inhibition [2-4] enhances the synaptic signal of photoreceptors, which is specified as a retinal Point Spread Function (PSF) seen as a biological convolution with the edge enhancement property [3]. At the second synaptic level, the lateral inhibition mediates the more complex properties such as the responses of directional selective Receptive Fields (RFs) [2].

The complexity of inter-neural circuitries, activations and responses of the retinal cells have been investigated [5, 6] in a search for the specific encoding role of each individual cell in the retinal processing leading to new insights. These includes the existence of a diverse range of retinal ganglion cells (RGCs) in which the size of each individual type varies in relation to the eccentricity of neurons and the distance from the fovea [5] supporting our biological understanding of the retinal multiscale encoding [7], completed in the cortex [5, 6, 8]. ON and OFF cells of each specific type are noted to have a variant size [5, 9] as well. It is also reported that there are different channels for passing the encoded information of ON-center and OFF-surround (and vice versa) activations of retinal RFs [6] to the cortex. Moreover, the possibility of simultaneous activations of a group of RGCs (as a combined activity) in the retina by the output of amacrine cells is noted in the literature [10-12]. Some retinal cells have been found with a directional selectivity property such as the cortical cells [5, 6]. It is noteworthy that despite the complexity and variety of retinal cells circuitry and coding, there are a few constancy factors common to them, valid even for amacrine and horizontal cells. The constancy of integrated sensitivity is one of these factors mentioned in the literature [13-15] which is quite useful for quantitative models for visual system.

The perception of directional tilt in the Café Wall illusion might tend to direct explanations toward the cortical orientation detectors or complex cells [8, 16]. We have shown that the emergence of tilt in the Café Wall illusion specifically [17-21], and in Tile illusions generally [17, 22, 23], is a result of simple cells processing with circularly symmetric activations/inhibitions. Low-level Filtering Models [24, 25], commonly apply a filter similar to a Gaussian or Laplacian of a specific size on the Café Wall to show the appearance of slanted line segments referred to as Twisted Cord [26] elements in the convolved output. These local tilts are assumed then to be integrated into continuous contours of alternating converging and diverging mortar lines at a more global level [22, 24, 25]. A hybrid retino-cortical explanation as a mid-level approach containing light spread, compressive nonlinearity, and center-surround transformation have been proposed by Westheimer [27]. Some other explanations relies on Irradiation Hypothesis [28] and Brightness Induction [29]. There are also high-level descriptive approaches such as 'Border Locking' [30] and 'Phenomenal Model' [31] for the illusion with little consideration to the underlying neurological mechanisms involved in the emergence of tilt in the Café Wall illusion.

Modelling the receptive field responses dates back to Kuffler's demonstration of roughly concentric excitatory center and inhibitory surround [32]. Then, Rodieck and Stone [33] and Enroth-Cugell and Robson [34] modelled the center and surround signals of the photoreceptors by two concentric Gaussians with different diameters. The computational modelling of early visual processing were followed by Marr and Ullman [35] who were inspired by Hubel and Wiesel's [8] discovery of cortical simple and complex cells. Laplacian of Gaussian (LoG) have been proposed by Marr and Hildreth [36] as an optimal operator for low-level retinal filtering and an approximation filter of difference of Gaussians (DoG) instead of LoG, considering a ratio of ~1.6 for the Gaussians diameters.

The model here [17-22] is a most primitive implementation for the contrast sensitivity of RGCs based on classical circular center and surround organization of the retinal RFs [33, 34]. The output of the model is a simulated result for the responses of the retinal/cortical simple cells to a stimuli/image. This image representation is referred to as an 'edge map' utilizing difference of Gaussians (DoG) at multiple scales to implement the center-surround activity as well as the multiscale property of the RGCs. Our explanation differs from the previous low-level models [24, 25, 28, 37] due to the concept of filtering at multiple scales in our model in which the scales are tuned to the resolutions of image features, not the resolutions of the individual retinal cells. We show also that our model is a quantitative approach capable of even predicting the strength of the Café Wall illusion based on different characteristics of the pattern [21].

This work is a complete collection of our findings on the underlying mechanism involved in our foveal and peripheral vision for modelling the perception of the induced tilt in the Café Wall illusion. It draws together and extends our previous studies on the foveal/local investigations of tilt on Café Wall illusion [18, 19], and extends our investigations for the peripheral/global analysis of the perceived tilt not in just one specific sample (to overcome the shortcomings of our previous studies [18, 19]), but for variations of different configurations modelling the Gestalt perception of tilt in the illusion.

In Section 2, we describe the characteristics of a simple classical model for simulating the responses of simple cells based on Difference of Gaussians (DoG), and utilize the model for explaining the Café Wall illusion qualitatively (2.2.1) and quantitatively (2.2.2). Afterwards in Section 3, the experimental results on variations of foveal sample sets are provided (3.1), followed by the report of quantitative tilt results for variations of different configurations of the Café Wall illusion with the same characteristics of mortar lines and tiles but with different arrangements of a whole pattern (3.2), which then completed by a thorough comparison of the local and global mean tilts of the pattern found by our simulations (3.3). We conclude by highlighting the advantages and disadvantages of the model in predicting the local and global tilt in the Café Wall pattern and proceed to outline a roadmap of our future work (Section 4).

## 2 MATERIALS AND METHODS

### 2.1 Formal Description and parameters

Applying a Gaussian filter on an image generates a blurred version of the image. In our DoG model, the difference of two Gaussians (DoG) filtered outputs generates one scale of the *edge map* representation. For a 2D signal such as image *I*, the *DoG* output, modeling the retinal GCs responses with the center surround organization is given by:

$$DoG_{\sigma, s\sigma}(x,y) = I \times 1/2\pi\sigma^2 \exp[-(x^2+y^2)/(2\sigma^2)] - I \times 1/2\pi(s\sigma)^2 \exp[-(x^2+y^2)/(2s^2\sigma^2)] \quad (1)$$

where *DoG* is the convolved filter output, $x$ and $y$ are the horizontal and vertical distances from the origin respectively and $\sigma$ is the scale of the center Gaussian ($\sigma = \sigma_c$). $s\sigma$ in Eq (1) indicates the scale of the surround Gaussian ($\sigma_s = s\sigma_c$), and $s$ is referred to as *Surround ratio* in our model as shown in Eq (2).

$$s = \sigma_{surround} / \sigma_{center} = \sigma_s / \sigma_c \qquad (2)$$

Increasing the value of *s*, results in a wider suppression effect from the surround region, although the height of the surround Gaussian declines (normalized Gaussians are used in our model). A broader range of *Surround ratios* from 1.4 to 8.0 have been tested with little difference to our results. We have considered another parameter in the model for the filter size referred to as *Window ratio* (*h*). To generate edge maps we have applied DoG filters within a window in which the value of both Gaussians are insignificant outside the window. The window size is determined based on the parameter *h* that determines how much of each Gaussian (center and surround) is included inside the DoG filter, and the scale of the center Gaussian ($\sigma_c$) such that:

$$WindowSize = h \times \sigma_c + 1 \qquad (3)$$

+1 as given in Eq (3) guaranties a symmetric DoG filter. In the experimental results the *Window ratio* (*h*) have been set to 8 to capture more than 95% of the surround Gaussians in the DoG convolved outputs.

### 2.2 Model and Image Processing Pipeline

An image processing pipeline has been used [18-21] here to extract edges and their angles of tilt (in the edge maps), as shown in Fig 1 for a crop section of a Café Wall pattern of size 2×4.5 tiles (the precise height is 2Tiles+Mortar=2*T*+*M*). In this research, we concentrate on the analysis of the induced tilt effect in the Café Wall illusion, to include the details of the parameters used in the simulations in order to quantify the tilt angle in this stimulus by modeling our foveal and peripheral vision.

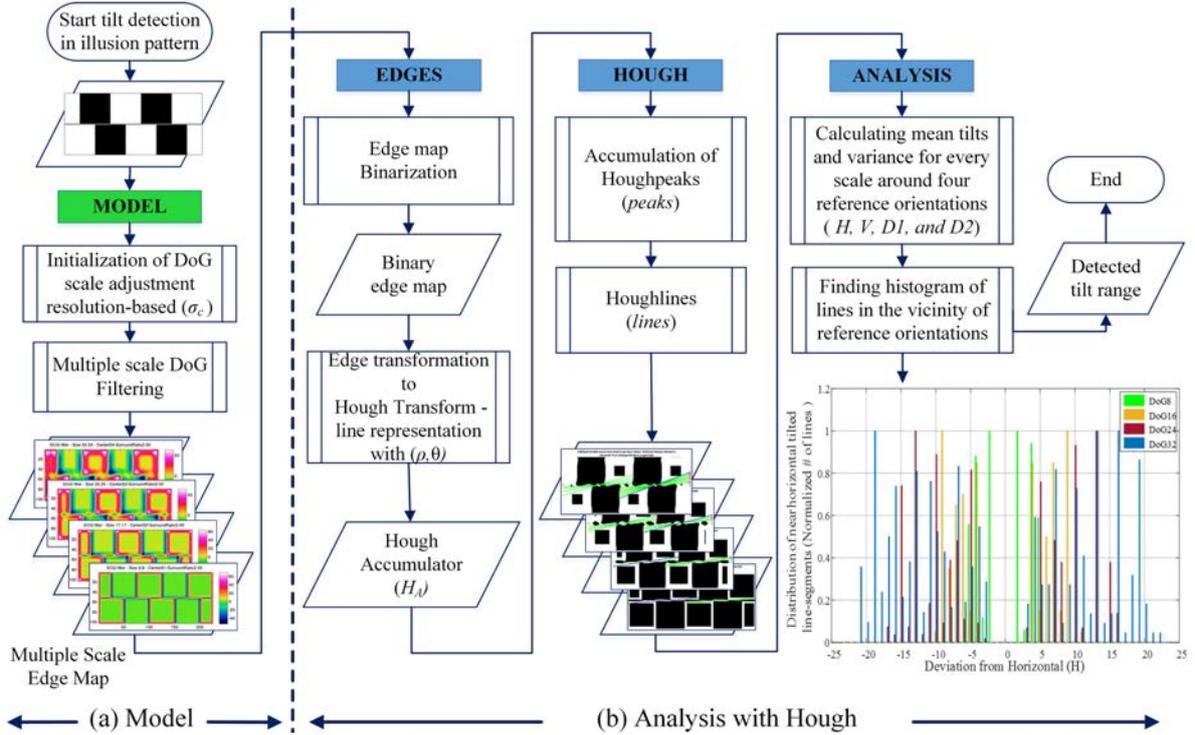

Fig 1. Flowchart of our model with Hough analytic processing pipeline (Reproduced with permission from [23]).

#### *2.2.1 DoG edge maps at multiple scales*

The DoG representation at multiple scales is the output of the model, which referred to as an *edge map* of an image. The DoG is highly sensitive to spots and moderately sensitive to lines that match the center diameter. We have used this representation for modelling the responses of visual simple cells especially on Tile illusions in our investigations [18-20, 22]. An appropriate range for $\sigma_c$ can be determined for any arbitrary pattern/image considering the pattern characteristics as well as the filter size matched with the image features (by applying Eq. 3) in our model. The step sizes determine the accuracy of the multiple scale representation here, and again are pattern specific for preserving the visual information with minimum redundancy but at multiple scales.

For Café Wall illusion, the DoG edge map indicates the emergence of divergence and convergence of the mortar lines in the pattern, similar to how it is perceived as shown in Fig 2. The edge map has been shown at six different scales in jetwhite color map [38] for a Café Wall of 3×8 tiles with 200×200px tiles (*T*) and 8px mortar (*M*). In order to extract the tilted line segments along the mortar lines referred to as Twisted Cord [26] elements, the DoG filter should be of the same size as the mortar size [18, 25, 37]. The edge map should contain both high frequency details as well as low frequency contents in the image. We start DoG filtering below the mortar size at scale 4 ($\sigma_c$=4; as the finest scale) and extend the scales gradually until scale 28 for a large filter to capture the tiles fully, with incremental steps of 4 (in the figure we have shown this till $\sigma_c$=24 due to shortage of space). Other noncritical parameters of the model are *s*=2 and *h*=8, representing the *Surround* and *Window ratios* respectively.

The DoG outputs in Fig 2 show that the tilt cues appear at fine to medium scales, and start to disappear as the scale of the center Gaussian increases in the model. At fine to medium scales of the edge map, there are some corner effects highlighting the appearance of tilted line segments result in the appearance of square tiles to look similar to trapezoids, which may be referred to as wedges in the literature [30], inducing convergent and divergent mortar lines. So, at fine scales around the size of the mortar, we see the groupings of identically coloured tiles with the Twisted Cord elements along the mortar lines. By increasing the scales gradually from the medium to coarse scales, when the mortar cues disappear completely in the edge map, other groupings of identically coloured tiles are emerged in the edge map, connecting tiles in zigzag vertical orientation. What we see across multiple scales in the edge map of the pattern are two incompatible groupings of pattern elements: groupings of tiles in two consecutive rows by the mortar lines at fine scales with nearly horizontal orientation (as focal/local view), and then groupings of tiles in zigzag vertical direction at coarse scales (as peripheral/global view). These two incompatible groupings occur simultaneously across multiple scales and exhibit systematic differences according to the size of the Gaussian and predicts the change in illusion effects with distance from the focal point in the pattern.

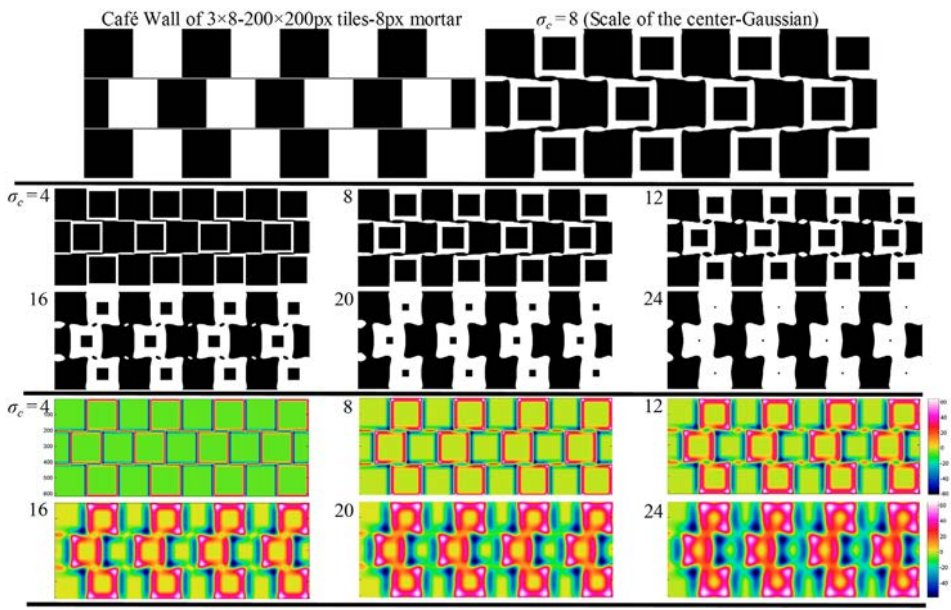

Fig 2. Top: Café Wall of 3×8 with 200×200px tiles and 8px mortar (left) and an enlarged DoG output at scale 8 ($\sigma_c$ =8) from the edge map of the pattern (right). Center: The binary edge map at six scales ($\sigma_c$=4 to 24 with incremental steps of 4). Bottom: The same edge map but presented in the jetwhite color map. The non-critical parameters of the DoG model are: *s*=2, and *h*=8 (*Surround* and *Window ratios* respectively; Reproduced with permission from [23]).

We have shown that in the edge map at multiple scales, not only the information of edges/textures are extracted with the shades and shadows around the edges, but also other cues related to tilt and perceptual grouping start to emerge as features for mid- to high-level processing [17, 22]. Also we have shown in another article that even the prediction of the strength of tilt effect in different variations of Café Wall illusion is possible from the persistence of mortar cues across multiple scales [21] in the edge map. Highly persistence mortar cues in the edge map is an indication for a stronger induced tilt in the stimulus.

### 2.2.2 Second stage processing

The Hough analysis is used for quantitative measurement of tilt in our model and consists of three stages of EDGES, HOUGH and ANALYSIS as shown in Fig 1, explained below.

*EDGES* – We used here an analysis pipeline to characterize the tilted line segments presented in the edge map of the Café Wall pattern. First, the edge map is binarized and then the Standard Hough Transform (SHT) [39, 40] is applied to it to detect line segments inside the binary edge map at multiple scales. SHT uses a two-dimensional array called the accumulator ($H_A$) to store line information of edges based on the quantized values of $\rho$ and $\theta$ in a pair ($\rho,\theta$) using hough function in MATLAB. $\rho$ specifies the distance between the line passing through the edge point and $\theta$ is the counter clockwise angle between the normal vector ($\rho$) and the x-axis ranges from 0 to $\pi$, $[0,\pi)$. Therefore, every edge pixel ($x,y$) in the image space, corresponds to a sinusoidal curve in Hough space such that $\rho=x.cos\theta+y.sin\theta$, with $\theta$ as free parameter corresponding to the angle of the lines passing through the point ($x,y$) in the image space. The output of *EDGES* is the accumulator matrix ($H_A$) with all the edge pixel information.

*HOUGH* – The *EDGES* provides all possible lines that could pass through every edge point of the edge map inside the $H_A$ matrix. We are more interested on detecting the induced tilt lines inside the Café Wall image. Two MATLAB functions called *houghpeaks* and *houghlines* are employed for the further processing of the accumulator matrix ($H_A$). The *houghpeaks* function finds the peaks in the $H_A$ matrix with three parameters of *NumPeaks* (maximum number of line segments to be detected), *Threshold* (threshold value for searching the peaks in the $H_A$), and *NHoodSize* (the size of the suppression neighbourhood that is set to zero after the peak is identified). The *houghlines* function extracts line segments associated with a particular bin in the accumulator matrix ($H_A$) with two parameters of *FillGap* (the distance between two line segments associated with the same Hough bin. Line segments with shorter gaps are merged into a single line segment), and *MinLength* (specifies keeping or discarding the merged lines. Lines shorter than this value are discarded).

A sample output of *HOUGH* processing stage is given in Fig 1 with the detected houghlines displayed in green on the binary edge map at four different scales (the crop section is selected from a Café Wall with 50×50px tiles and 2px mortar, with the DoG scales from 0.5$M$ to 2$M$ in the figure around the mortar size with the incremental steps of 0.5$M$). The results of *HOUGH* analysis stage for a different crop section of a Café Wall pattern with higher resolution (cropped from a Café Wall with 800×800px tiles and 32px mortar) are shown in Fig 3 for two scales of the DoG edge map ($\sigma_c$ = 32, 64 -$M$ and 2$M$; Blue lines indicate the longest line segments detected). The histograms of detected houghlines near the horizontal have been provided for these scales. The absolute mean tilts and the standard deviation of tilts are calculated and presented in the figure below the graphs.

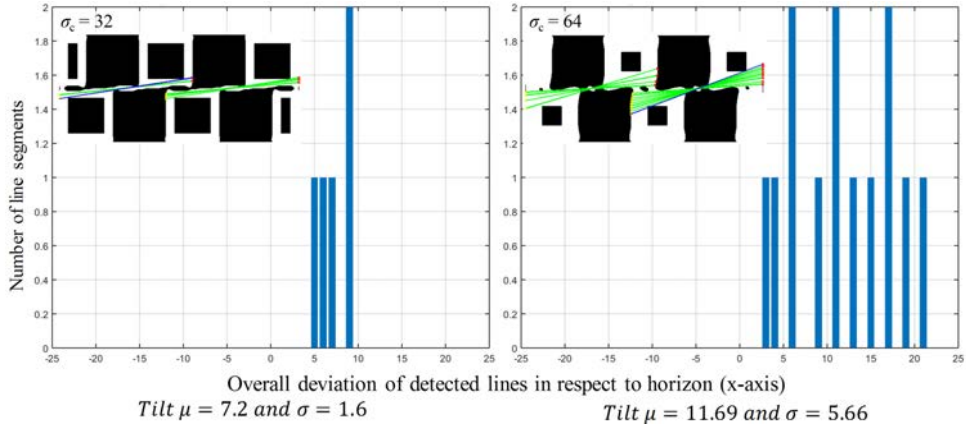

Fig 3. *HOUGH* stage output – Distribution of line segments detected near the horizontal, presented for two scales of the DoG edge map at scales 32 and 64 ($\sigma_c$ =32, 64). Mean tilt and variance for each graph have been also provided.

*ANALYSIS* – The detected line segments and their angular positions are saved inside four orientation matrices considering the closest to any of the reference orientations of Horizontal (*H*), Vertical (*V*), positive Diagonal (+45°,*D1*), and negative Diagonal (-45°, *D2*). We consider an interval of [-22.5°, 22.5°) around each reference orientation to cover the whole space. The statistical analysis of tilt angles of the detected lines around each reference orientation is the output of this stage and includes the mean tilts and the standard errors around the means for each scale of the DoG edge map.

***Hough Parameters for Tilt Investigations of Café Wall Stimulus*** – Recall that *NumPeaks* indicates the maximum number of line segments to be detected, *FillGap* shows the distance between two line segments associated with the same Hough bin in which line segments with shorter gaps are merged into a single line segment. The other Hough parameter is *MinLength,* specifies keeping or discarding of the line segments considering this minimum length, and discarding the lines shorter than this value. To select an appropriate value for these parameters we should consider pattern features, and the scales of the DoG edge map. In the Café Wall pattern, in order to detect the Twisted Cord elements at fine scales, the *MinLength* value should be in a reliable range. The Twisted Cord elements have a minimum length of 2.5$T$ (*MinLength*≈2.5$T$), and therefore for a Café Wall with 200×200px tiles, *MinLength*=500px. We set this parameter a bit smaller than this value equal to *MinLength*≈2.25$T$=450px for our experiments in Section 3. The *FillGap* parameter is chosen equal to 1/5[th] of a tile size (1/5$T$) in our experiments (to merge the disconnected mortar cues of each Twisted Cord elements at fine to medium scales in the edge maps). *NumPeaks* is selected appropriately based on the size of the pattern, and for small foveal sets (Section 3.1), this is set to 100 but has a higher range, 520 and 1000 for larger Café Wall stimulus for global investigation of tilt (Section 3.2).

# 3 RESULTS AND DISCUSSIONS

## 3.1 Local tilt investigation

### *3.1.1 Falling and Rising mortar investigation*

This work draws together and extends our previous studies on the foveal/local investigations of tilt on Café Wall illusion [18, 19], and the extension of our investigations for the peripheral/global analysis of the perceived tilt in this work not in just one specific sample but for variations of different configurations. The quantitative mean tilts of similar shape samples but with variant resolutions have been investigated in our previous work [20]. We have shown that for variations with different resolutions, the tilt prediction of the model stays nearly the same when the dependent parameters of the model to the spatial content of the pattern have been updated accordingly in each resolution ($\sigma_c$ and Hough parameters).

We report here the evaluation results of our model's predictions for the *direction of detected tilts* for two types of mortar lines in the Café Wall illusion [20]. Instead of referring to the mortar lines as either convergent or divergent, we rather talk about Falling or Rising mortar lines, in which in the Falling mortar, the direction of induced tilt is downwards on its right side compared to the horizontal direction and for the Rising mortar the vice versa. For instance in Fig 2 (top-left) the top mortar line is Falling while the bottom one is Rising. In this experiment the cropped samples specifically selected in such a way to contain only one mortar line indicates the emergence of tilt in only one direction of either positive or negative in the DoG edge maps (Falling or Rising). The samples have a height of two tiles and the mortar line in between ($2T + M$), and the width of 4.5 tiles ($4.5T$; $T$: tile size, $M$: mortar size), with the same height above and below the mortar line. In Section 3.1.2, we show the results for samples of variant sizes and different cropping technique for a more general investigation of the foveal/local perception of the inducing tilt in the pattern.

To fix parameters not being investigated, we restrict consideration initially to the Café Wall of 3×9 tiles with 400×400px tiles and 16px mortar. Here, in a systematic approach, 50 samples were selected from the Falling mortar and 50 samples from the Rising mortar with the dimensions described above from the Café Wall of 3×9 tiles. The sampling process starts from the left most side of the pattern, and with a horizontal shift size/offset of 32 pixels between the samples for the cropping window. A few examples of the Falling and Rising samples have been provided in Fig 4. The cropped samples at the bottom of the figure are symmetrical crops from the Rising mortar lines from the stimulus (Café Wall of 3×8). In the DoG edge maps of these samples, the scale of the center Gaussian is in the range of $1/2M$ to $2M$ with the incremental steps of $1/2M = 8$px ($\sigma_c$ = 8, 16, 24, 32) to detect both mortar lines and the outlines of the tiles for detecting near horizontal tilts in the edge maps.

For individual samples of the Falling and Rising mortar, the near horizontal mean tilts and variance of the detected houghlines have been shown in Fig 5. As the scale of the center Gaussian ($\sigma_c$) in our model increases, the variance of tilt also increases. The mean tilt results of the Falling and Rising mortar samples in Fig 5 indicates that both types of mortar lines follow nearly the same pattern.

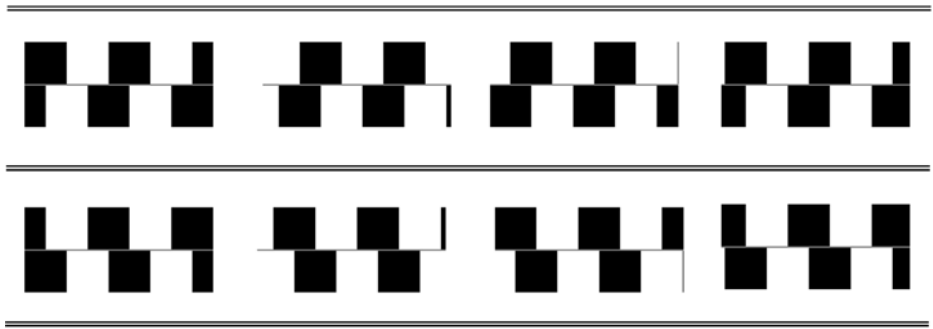

Fig 4. Examples of Systematic cropped samples along Falling (Top) and Rising mortar lines (Bottom), selected from Café Wall of 3×9 tiles with 400×400px tiles (*T*) and 16px mortar (*M*) - Café Wall 3×9 T400-M16. In total, 50 samples are taken along each mortar line with an offset of 32px between samples in each step. Crop samples have a size of $(2T+M)×4.5T$ with the mortar line positioned in the middle.

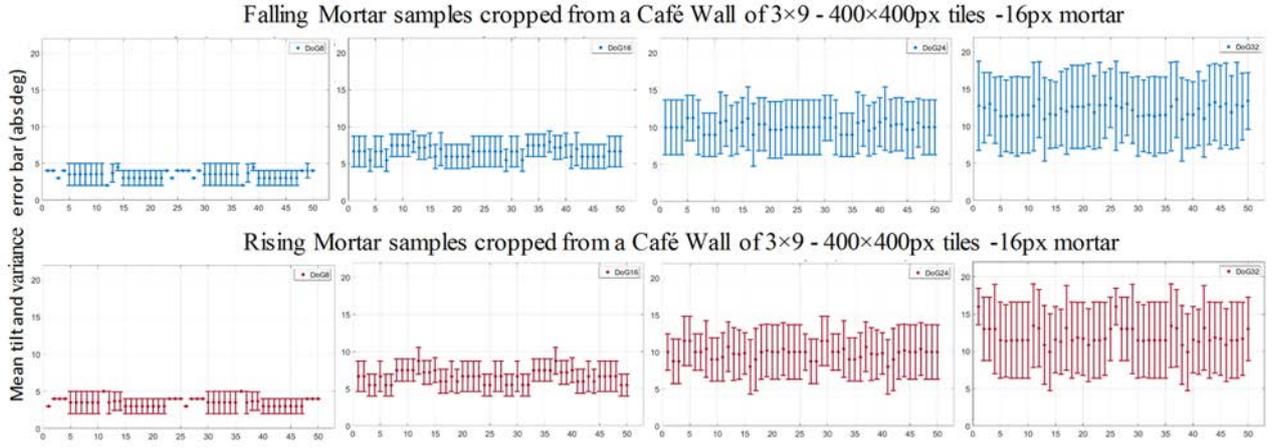

Fig 5. Mean tilts and variance error bars for individual samples of Falling (Top) and Rising mortar lines (Bottom) specified along horizontal axis (100 samples in total). As explained in Fig 4, the crop samples are from the Café Wall of 3×9 tiles with 400×400px tiles and 16px mortar (Reproduced with permission from [23]).

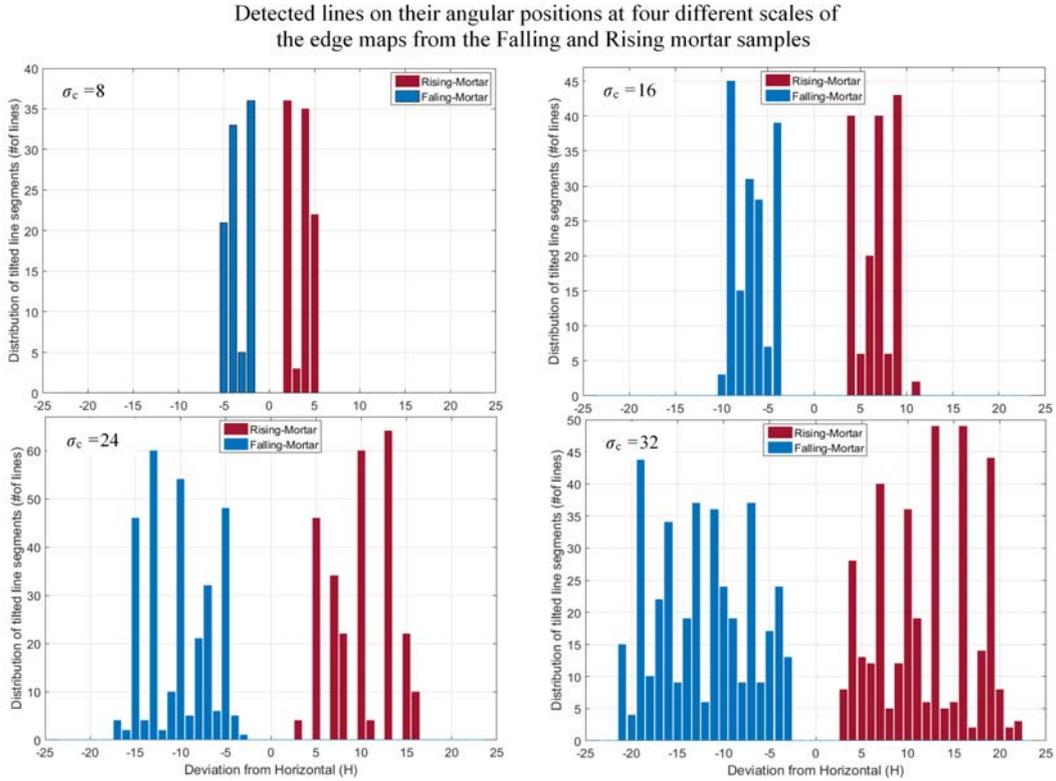

Fig 6. Distribution of line segments detected by deviation in degrees from the horizontal (x-axis), for the total 100 samples of Falling and Rising mortar lines, as explained in Fig 4. The scales of the edge maps ($\sigma_c$) ranges from 8 to 32 with step sizes of 8 ($1/2M$ to $2M$ with incremental steps of $1/2M$=8px). Other parameters of the model and Hough analysis are: $s$=2, $h$=8 (the *Surround* and *Window* ratios), with *NumPeaks*=50, *Threshold*=3, *FillGap*=80, and *MinLength*=960 as the Hough parameters (see [20] for the resolution analysis and its effect on the Hough parameters; Reproduced with permission from [23]).

The near horizontal line segments detected in the DoG edge maps (houghlines) at four scales ($\sigma_c$ = 8, 16, 24, 32) are shown in Fig 6 for the Falling (Blue bars) and the Rising (Red bars) mortar samples. These graphs are summarized in Fig 7 in a single graph, indicating normalized distribution of line segments detected with their deviations in degrees from the horizontal (x-axis) for the 100 samples. When the DoG scale increases, the detected tilt range covers a wider neighbourhood area around the horizontal axis as their details given in Fig 6. The deviations of the detected lines from the horizontal in Figs 6 and 7 are very small at the finest scale ($\sigma_c$ = 8). The range of tilt angles increases along the following scales of the edge maps reaching to a wide range of variations at scale 32 ($\sigma_c$ = 32 or DoG32) that is not reflected in our subjective experience of tilt in the pattern (it is overestimated at this scale). In the literature [18, 19, 22, 25, 37] it is noted that the size of DoG filter should be close to the size of the mortar for the Twisted Cord elements to be appeared along the mortar lines, here DoG16 ($\sigma_c$ =16). We have demonstrated that this is not applicable for Café Wall patterns with very thick mortar lines [21]. In this case, the mortar cues are lost completely in the edge map even by applying DoG filters smaller than the mortar size. The strength of the illusion are highly dependent on the characteristics of the pattern such as the luminance of the mortar, the mortar size, the contrast of the tiles, the aspect ratio of the tile size to the mortar size, and other parameters of the stimulus. We noted that there exists a correlation between the strength of the illusion with the persistence of the mortar cues in the edge maps across multiple scales [21].

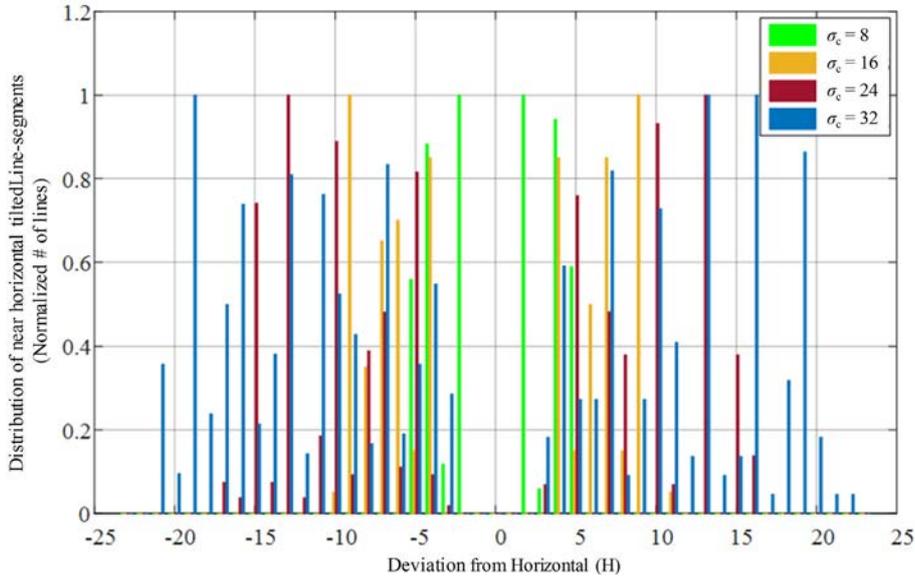

Fig 7. The normalized graph of lines detected from the DoG edge maps of the 100 cropped samples by deviation in degrees from the horizontal (H). Samples are from the Falling and Rising mortar lines from the Café Wall 3×9-T400-M16. The characteristics of the samples are provided in Fig 4, and the edge maps have four different scales ($\sigma_c$ = 8, 16, 24 and 32; Based on [20]).

### 3.1.2 *Variant sampling sizes – two methods of sampling*

As explained in Section 2.2.2, an analytic processing pipeline is used to quantify the tilt angles in the DoG edge maps. For modelling variant foveal views, several sampling sizes and aspect ratios have been investigated across multiple scales in order to find the confidence intervals around the predicted tilts reported in our previous work [18, 19]. These variations are verified and quantified in simulations using two different sampling approaches. The contrast of local tilt detection with a global average across the whole Café Wall pattern will be discussed in Section 3.2.

The eyes process the visual scene at different scales at different times, due to the intentional and unintentional eye movements while we look at the scene (pattern). Notably overt saccades and gaze shifts result in a rapid scan of the field of view by the fovea for the pertinent high-resolution information. Our visual perception of tilt in the Café Wall is affected by our *fixation* on the pattern. The induced tilts weakened in a region around fixation point, but the peripheral tilts stay unaffected with stronger tilts. It seems that in the Café Wall illusion, the final induced tilts gets greater effect from the peripheral tilt recognition compared to the foveal/local tilt perception. The possible correlations that might exist between the tilt effect to our foveal/peripheral view of the pattern due to gaze shifts and saccades are further investigated here. The local 'cropped' samples simulate foveal-sized locus only, but different scales of the DoG edge maps represent different degrees of eccentricity (the distance from the fovea) in the periphery.

In this section we are reporting the experimental results from [18, 19] and we restrict consideration initially to Café Wall of 9×14 tiles with 200×200px tiles and 8px mortar (Fig 8-left), with three "foveal" crop sizes to be explored, Crop4×5 (Crop section of a 4×5 tiles), Crop5×5, and Crop5×6 (Fig 8-right). Although the size of foveal image can be estimated by factors such as specific image size, viewing distance or human subject in mind (which usually are considered in psychophysical experiments), but the sample sizes explored in our experiments for the simulation results are selected for convenience without considering those restrictive factors.

Two sampling methods have been applied: *Systematic* and *Random cropping*. In the '*Systematic Cropping*' [18] for each specified crop window size, 50 samples are taken from the Café Wall of 9×14 tiles, in which the top left corner for the first sample is selected randomly from the pattern and for the rest of the samples, the cropping window shifts horizontally to the right with an offset of 4 pixels in each step. The total shift is equal to a tile size (200px) at the end, so, there is no repeated versions of any samples. In the '*Random Cropping*' [19] approach, for each specified crop window size, 50 samples are taken from randomly selected locations with the only consideration of the crop borders to stay inside the pattern.

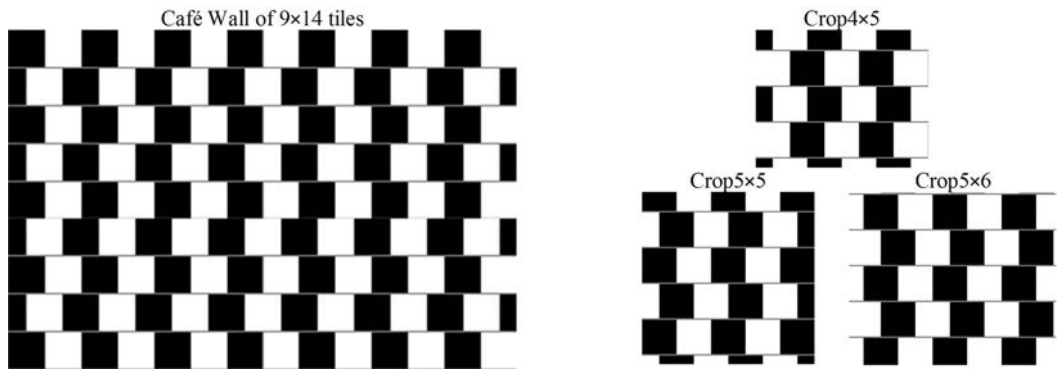

Fig 8. Café Wall of 9×14 tiles with 200×200px tiles and 8px mortar (Left), and three 'foveal' sample sizes explored (Based on [18, 19]). (CropH×W is H×W tiles).

The range of DoG scales of these samples are from $0.5M$ to $3.5M$ with the incremental steps of $0.5M$, and coarser scales exceeding the tile size ($T$; using Eq (3)), result in a very distorted edge pattern. We calculated [19] not only the near horizontal mean tilts, but also the vertical and diagonal tilts at medium to coarse scales in this experiment. Unlike the previous experiment in Section 3.1.1, detecting only horizontal tilts, we extended the range of scales in the models from $2M$ in the previous experiment to $3.5M$ for these experiments. With DoG filters larger than the mortar size, ultimately reaching to the coarsest size selected (3.5M, using Eq. 3), then the tiles are fully captured at the coarsest scales of the edge maps. These are being used for the vertical and diagonal deviations and the groupings of tiles in zigzag vertical orientation in our investigations. The Hough parameters (of both houghpeaks and houghlines functions) should have a proper range to detect the near horizontal slanted line segments in the edge maps at fine scales (refer to Section 2.2.2). For example, *FillGap* should assign a value to fill small gaps between the line segments appeared in the edge map at fine scales to detect near horizontal tilted lines, and *MinLength* should be larger than an individual tile size ($T$) to avoid the detection of the outlines of the tiles in the calculations. The value of Hough parameters depend on the pattern's attributes (features) and they are selected empirically for the tilt investigation in the experiments. To attain reliable and comparable tilt results, a constant set of these parameters have been used in this experiment and in Section 3.2 for the global tilt investigation, which is: *NumPeaks*=100, *Threshold*=3, *FillGap*=40, and *MinLength*=450. Other values for *NumPeaks* have been also tested for the global tilt investigation in Section 3.2 (520 and 1000).

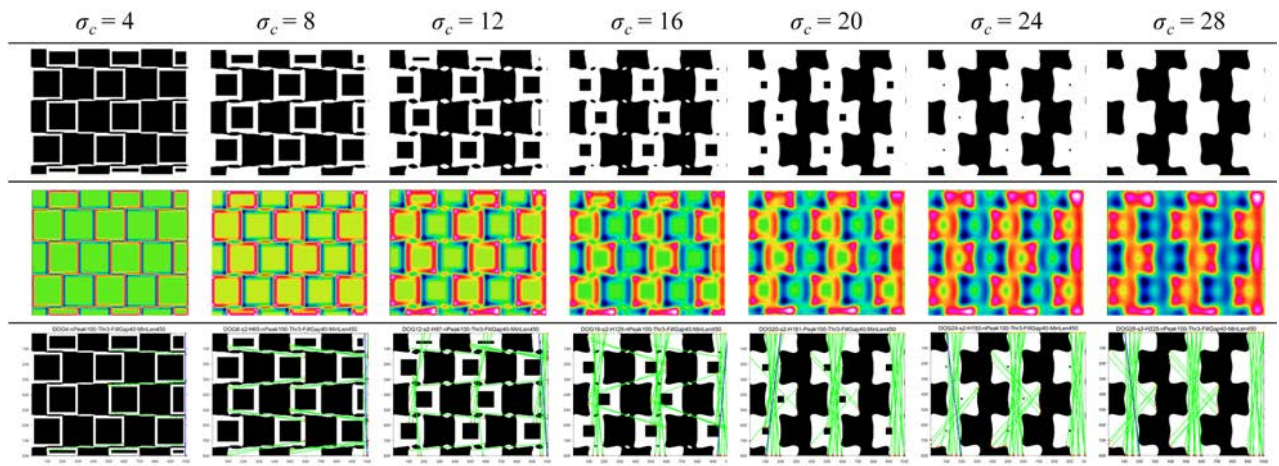

Fig 9 Top: A binary DoG edge map at seven scales ($\sigma_c$ = 4, 8, 12, 16, 20, 24, 28) of a crop section of size 4×5 tiles from a Café Wall with 200×200px tiles and 8px mortar. Middle: The DoG edge map is displayed in jetwhite colormap. Bottom: Detected houghlines from the edge map displayed in Green, overlayed on the binary edge map with Hough parameters as: *NumPeaks*=100, *Threshold*=3, *FillGap*=40, and *MinLength*=450 (Reproduced with permission from [23]; based on [19]).

Fig 9 shows a binary DoG edge map at seven different scales (top), the edge map presented in the jetwhite color map (middle), as well as the detected houghlines displayed in green on the edge map (bottom) for a sample of a Crop4×5 tiles, selected from the Café Wall of 9×14 tiles (Fig 8).

The absolute mean tilts in box plot have been graphed for the detected lines in the DoG edge maps at seven different scales for each sample set, the two sampling methods and around the four reference orientations of Horizontal (*H*), Vertical (*V*), and Diagonals (*D1, D2*) in Fig 10 (a,b). As the figure indicates, the *'Random Cropping'* approach produces more stable tilt results across variant foveal sample sizes compared to the *'Systematic'* sampling method. We noted [17] that the *Systematic* sampling approach is closer to the bias of our saccades and gaze shifts toward interest points, but the *Random* Sampling is a more standard statistical approach. At fine to medium scales of both sampling methods, there are only horizontal and vertical lines detected. A few samples of Crop5×5 have *D2* components at scale 16 due to border effects-only 4/50 samples). The results for near horizontal mean tilts at scale 8 ($\sigma_c$ = 8) show a nearly stable range around 7° in all samples (the DoG filter size apparently correlates with the mortar size). As the scale increases from 20 onwards, there are no near horizontal lines detected, but more vertical and diagonal lines are extracted from the edge maps. This is because the mortar cues in the edge maps at these scales start to fade, and also the enlargement of the outlines of the tiles results in more lines detected around the vertical and diagonal orientations at coarse scales of the DoGs. By increasing the scale, the horizontal mean tilts also increase and at scales 8 and 12 it is nearly 8°, however, at the finest scale ($\sigma_c$ = 4) the horizontal mean tilt is quite small (~3.5°). When we fixate on the pattern, we encounter with a weaker tilt effect, since similarly in the fovea the acuity is high because of high density of small size receptors. The vertical mean tilts are approximately 5-6°at medium to coarse scales. The diagonal mean tilts (around *D1* and *D2* axis) are around 4-5° which can be seen again at medium to coarse scales ($\sigma_c$ = 20, 24, 28).

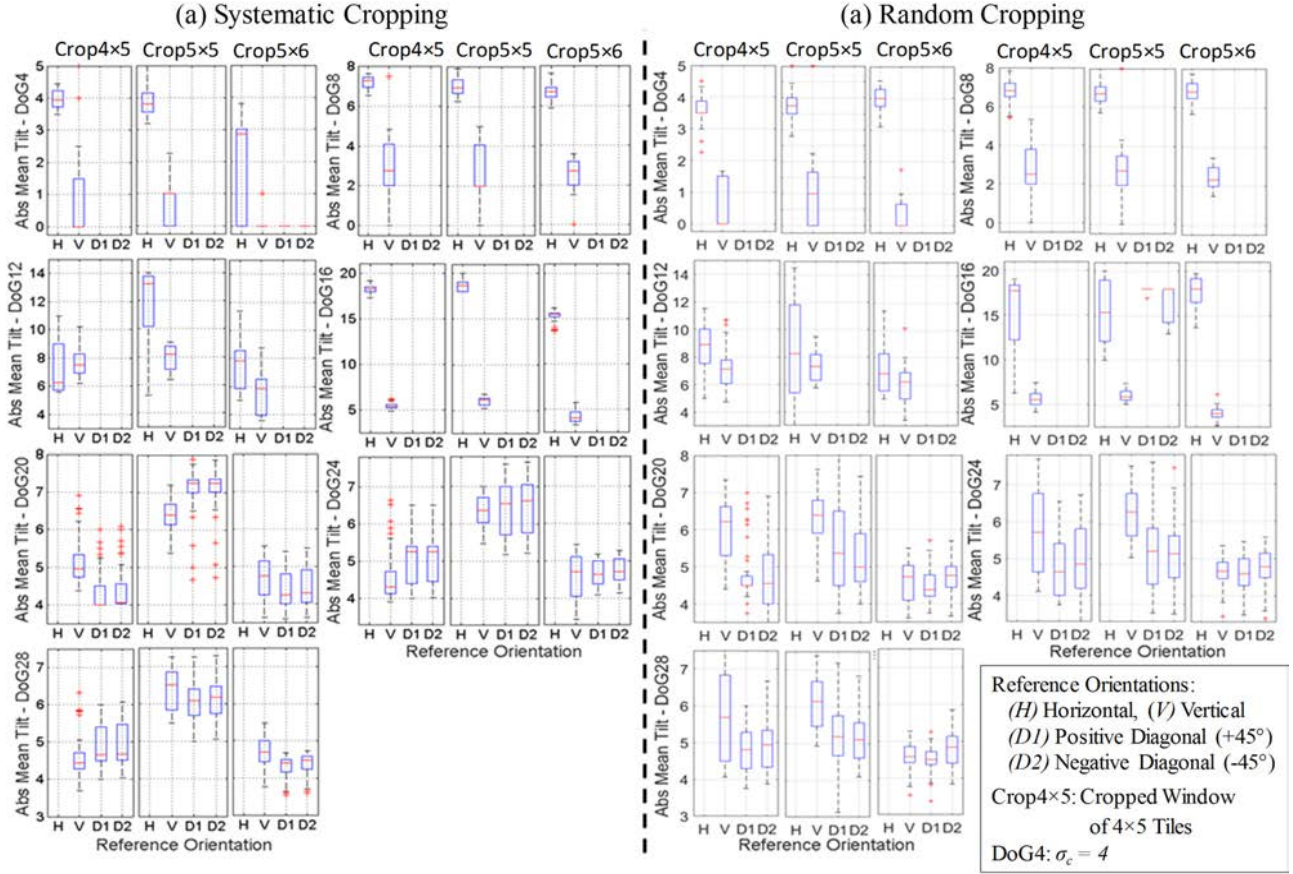

Fig 10. Mean tilts and standard errors around the four reference orientations (*H, V, D1, D2*), for the three 'foveal' sample sets (Fig. 8) with the two sampling approaches of (a) Systematic (Left) and (b) Random (Right) methods. The parameters of the model and analytic Hough processing are: edge maps at seven DoG scales ($\sigma_c$ =4, 8, 12, 16, 20, 24, 28), *s*=2, *h*=8, with Hough parameters of *NumPeaks*=100, *Threshold*=3, *FillGap*=40, and *MinLength*=450 (Reproduced with permission from [23]; based on [19]).

The results of the detected mean tilts at a given scale shows slight differences across foveal sample sets, and this is expected because of the random sampling and the fixed Hough parameters that are not optimized for each scale and sampling size, and are kept constant here for the consistency of the higher level analysis/model. The tilt detection results are sensible when compared to our angular tilt perception of the pattern, but more accuracy may be achieved by optimizing parameters based on the psychophysical experiments.

Fig 11 (a, b) shows the distribution of lines detected from the DoG edge maps at seven scales, and around the four reference orientations (*H, V, D1, D2*) for the three foveal sample sets and the two sampling methods. The near diagonal tilted lines (around *D1* and *D2* axes) have been graphed together for fairer representation. Fig 11 (a, b) shows that the houghlines detected in (b) is more normally distributed around the reference orientations compared to (a). All the graphs indicate the effect of the edge maps at multiple scale on the range of detected mean tilts and the distribution of the lines detected by their deviations in degrees from the reference orientations. The mean tilts covers a wider angular range when the DoG scale increases. We should remind that the number of detected lines is highly dependent with the sample size and the *NumPeaks* parameter. We explain the tilt results in Fig 11 (b) but the same explanation can be used for part (a).

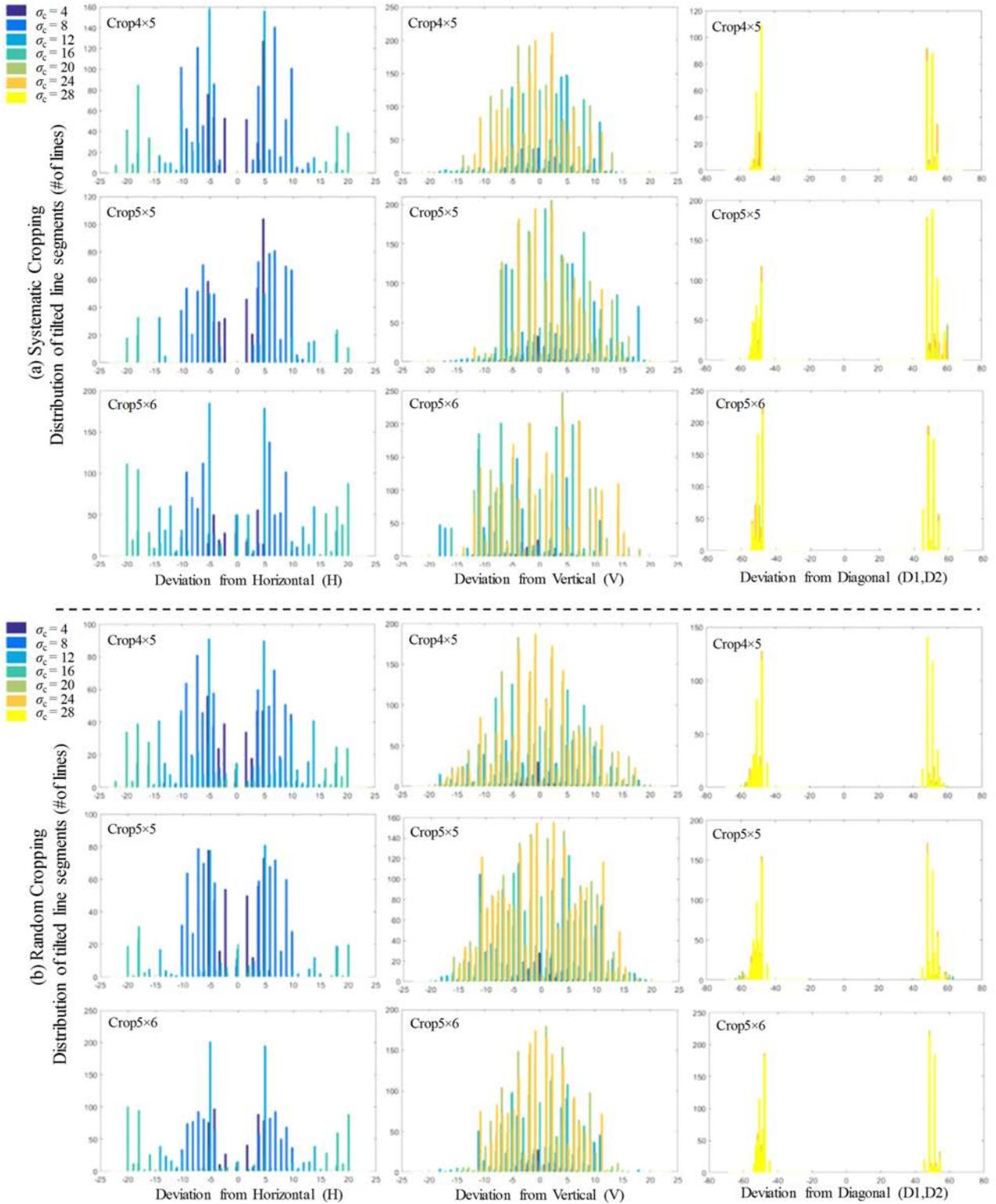

Fig 11. The distribution of the line segments detected from the edge maps of the foveal sets (Fig. 8), having either horizontal (Left-column), vertical (Center-column), or diagonal (Right-column) deviations, with the two sampling methods: (a) Systematic (Top), and (b) Random cropping (Bottom). The edge maps are at seven different scales ($\sigma_c$ =4, 8, 12, 16, 20, 24, 28) and a fixed set of Hough parameters are used as *NumPeaks*=100, *Threshold*=3, *FillGap*=40, and *MinLength*=450 (Reproduced with permission from [23]; based on [19]).

Fig 11 (b) (left-column) shows the near horizontal lines detected for the three foveal sets. At scale 4 ($\sigma_c$ = 4), the detected tilt angles are very small, ranging between 2-5°, with the peak of 4°. Furthermore, at scale 16, the detection of a high range of variations of tilt angle is not reflected in our perception of the pattern. To detect horizontal tilt cues along the mortar lines, the scale of the center Gaussian in our model should be close to the mortar. At scale 8, the mean tilts are between 3-10° with the peak of 7° for most lines, and at scale 12 it is increased to ~14°. At scale 16 the results show a wider range of horizontal lines detected and a fairly broad range of vertical lines, and this fits as a transition stage between the 'horizontal groupings' of identically colored tiles with the mortar lines at a focal view, and the 'zigzag vertical groupings' of the tiles [22] at a more peripheral/global view.

Fig 11 (b) (center-column) shows the near vertical lines detected. Similar to the indications of Fig 10, they start to be detected at fine scales due to some edge effects in a few samples, but as color code indicates, the majority of the near vertical lines are detected at scales 20 and 24, with the mean tilts in the range of 2-15° and the peak close to the *V* axis. In Fig 11 (b) (right-column) the detected lines with near diagonal deviations indicate that the dominant scales for detection of the diagonal lines are mainly at coarse scales of 24 and 28 ($\sigma_c$ =24, 28) with approximately 1-2.5° deviation from the diagonal axes (*D1*, *D2*).

### 3.2 Global tilt investigation

#### 3.2.1 *Global Tilts in the Café Wall of 9×14 tiles*

The Café Wall illusion is characterized by the appearance of Twisted Cord elements along the mortar lines [24-26], making the tiles seem wedge-shaped [30]. These local tilt elements are believed to be integrated and produce slanted continuous contours along the whole mortar lines by the cortical cells [16, 28] result in alternating converging and diverging mortar lines at a global view.

Because the tilt effect in the Café Wall is highly directional, it raises the question of whether lateral inhibition and point spread function (PSF) of retinal cells can explain the tilt effect in the pattern or not. We demonstrated that a bioplausible model [17-22], with a circularly symmetric organization as a simplified model for the retinal GC responses [33, 34], is able to reveal the tilt cues in the Café Wall illusion across multiple scales of the edge map. To explain the emergence of tilt in the Café Wall, there is no need to utilize complex models of non-CRFs [41-45] implementing the retinal/cortical orientation selective cells.

In this experiment, the intension is to investigate the Gestalt pattern, simulating peripheral awareness across the entire image and overcome the shortcomings of our previous investigations. We investigate here the global tilts in the Café Wall of 9×14 with 200×200px tiles and 8px mortar (Fig 8-left) and on its DoG edge map at seven different scales to quantify the tilt angles around the four reference orientations. The DoG scales have a range from $0.5M$ to $3.5M$ with the incremental steps of $0.5M$ the same as the foveal samples in Section 3.1.2.

In our first attempt to examine the robustness of the model for global tilt analysis [18, 19], the analysis was done with the parameters appropriate for local features. We have tested *NumPeaks*=100 in [18] and *NumPeaks*=520 in [19], but we have not achieved convincing results. Increasing the value of *NumPeaks* from 100 to 520 did not show any significant change to the mean tilts although it substantially increased the variance. The results showed that the near horizontal mean tilts was approximately 7° at scale 8 nearly the same as the foveal sample sets (Fig 10). The near vertical mean tilts at medium to coarse scales were around 2°, while they were around 6° in the foveal sets. The near diagonal mean tilts were approximately 3° and they were in the range of 5-6° in the foveal sample sets. Please refer to [19: Fig 6] for more details. In this work, we perform a global analysis with larger numbers of line segments as appropriate to the large global pattern. *NumPeaks* is a size relevant parameter, and its value is critical for achieving reliable results. Increasing this value for hough analysis on the foveal sets does not affect on the detected houghlines there, but an appropriate value for large samples are essential to detect all the relevant houghlines available in the edge map with smooth variations reflecting our estimation of tilt that is comparable with the detected lines in the foveal sets in our simulations.

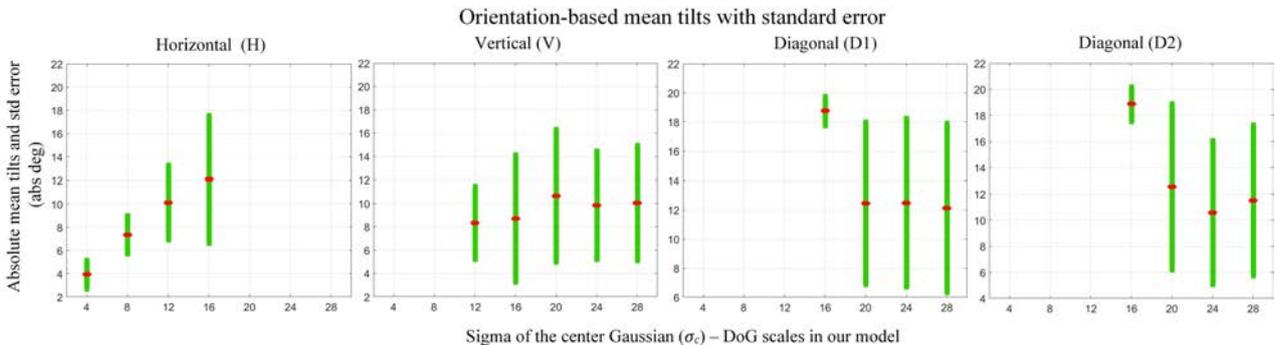

Fig 12 Mean tilts and the standard errors of detected tilt angles around the four reference orientations (*H, V, D1, D2*) from the DoG edge map at seven different scales of the whole Café Wall 9×14 with 200×200px tiles and 8px mortar. Green errorbars correspond to Hough *Num-Peaks*=1000, with mean values shown in Red.

The new experimental results for mean tilts and standard errors of the detected tilt angles have been presented in Fig 12 for the Café Wall of 9×14 tiles with *NumPeaks*=1000. The other parameters are kept the same as Figs 10 and 11 for the foveal sets. As indicated in the left graph in Fig 12, the near horizontal mean tilt is approximately 7.5° at scale 8 nearly the same as the foveal sample sets (Fig 10). In the horizontal graph, we see that by increasing the DoG scale, the mean tilt also increases from 7.5° to ~10° at scale 12 with higher variations compared to the foveal sample sets. The new results for the vertical and diagonal mean tilts at coarse scales have been improved dramatically from our previous reports (explained in previous paragraph), and show a variation of tilt angles for the detected houghlines. The near vertical mean tilts at medium to coarse scales were around 2° that are quite negligible in the previous report; now they are around 10° while they were around 6° in the foveal sets. The near diagonal mean tilts at medium to coarse scales were approximately 3° in the previous reports; now it is ~10° and they were in the range of 5-6° in the foveal sample sets. We have shown here that the new results with periphery appropriate parameterization are reliable and comparable with the previous results for foveal parameterization (Section 3.1.2). We will explain more on these results in Sections 3.2.2 and 3.3.

### 3.2.2 Variant Sized Café Wall Patterns with the Same Aspect Ratios of Tile size to Mortar Size

We can assume that the tilt perception of the Café Wall illusion starts by a wholistic view to the pattern, which then extends to a local focusing view along the mortar lines in search for further cues of tilt in the pattern. Both of these local and global views to the Café Wall have their own effect on the strength of our perceptual understanding of tilt in this pattern. We started our investigations of the global tilt analysis in the Café Wall stimulus by first addressing the shortcomings of our previously reports [18, 19] as reflected in the previous section and presented reliable tilt results for a specific sample (Café Wall of 9×14 tiles) based on periphery appropriate parameterization. We show in this experiment our deep investigations of the global tilts on variations of Café Wall with the same characteristics of mortar lines and tiles but with different arrangements of a whole pattern. We have explored here the correlation between the tilt effect and the layout of the pattern in general (how the tiles are arranged to build the stimulus).

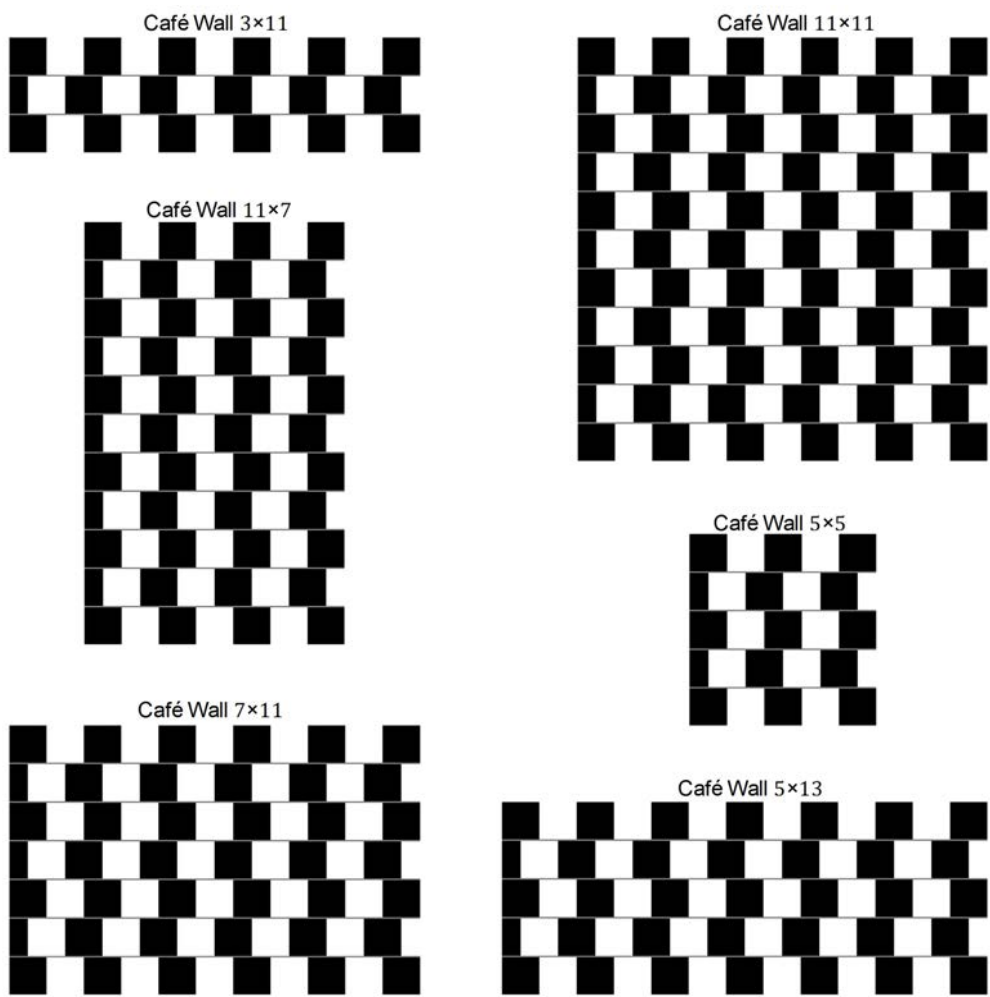

Fig 13 Different configurations of Café Wall pattern, from Café Wall of 3×11 tiles on Top-left to Café Wall of 5×13 tiles at the Bottom-right of the figure. All the patterns have the same tile size of 200×200px and the mortar size of 8px (Reproduced with permission from [23]).

In this experiment, variations of Café Wall pattern have been investigated with the same aspect ratios of tile size ($T$) to the mortar size ($M$) ($T/M=const.$) in order to check whether #rows and #columns in the overall arrangement of tiles in the Café Wall pattern have an effect on the detected tilts in our simulation results or not. In other words, we check the Gestalt perception of the Café Wall pattern, and its relation to visual angle of the whole pattern (not just the visual angle of an individual tile and mortar line investigated so far [18, 19, 21]). We show here that our model can predict a slightly different tilt results for these variations, similar to our global perception of illusory tilt in the pattern in the same way as human is affected by the configurations of the pattern. This is being reported for the first time with the quantitative results.

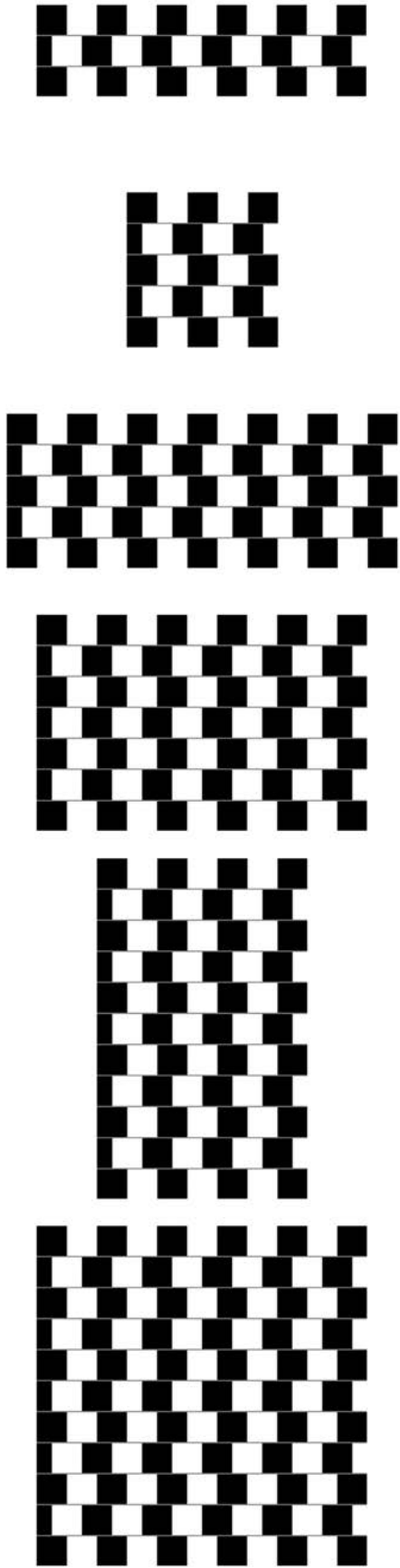

| Café Wall 3×11 (Mean Tilt ± StdErr) | | | | |
|---|---|---|---|---|
| $\sigma_c$ | H | V | D1 | D2 |
| 4 | 04.50±0.25 | NaN | NaN | NaN |
| 8 | 07.15±0.54 | 14.00±0.0 | NaN | NaN |
| 12 | 10.00±0.59 | 10.63±0.49 | NaN | NaN |
| 16 | 12.74±1.07 | 07.30±0.58 | 18.71±0.72 | 19.40±0.45 |
| 20 | NaN | 07.20±0.66 | 12.84±0.84 | 12.45±0.82 |
| 24 | NaN | 07.13±0.64 | 13.06±0.77 | 12.93±0.70 |
| 28 | NaN | 07.55±0.59 | 13.69±0.77 | 14.07±0.72 |

| Café Wall 5×5 (Mean Tilt ± StdErr) | | | | |
|---|---|---|---|---|
| $\sigma_c$ | H | V | D1 | D2 |
| 4 | 04.00±0.0 | NaN | NaN | NaN |
| 8 | 07.33±0.63 | NaN | NaN | NaN |
| 12 | 09.46±1.01 | 08.66±0.60 | NaN | NaN |
| 16 | 12.94±1.04 | 08.02±0.70 | 19.33±0.45 | 19.00±0.57 |
| 20 | NaN | 09.78±0.72 | 13.11±0.89 | 12.89±0.90 |
| 24 | NaN | 10.14±0.63 | 12.08±0.70 | 12.37±0.68 |
| 28 | NaN | 10.82±0.68 | 12.54±0.72 | 12.74±0.69 |

| Café Wall 5×13 (Mean Tilt ± StdErr) | | | | |
|---|---|---|---|---|
| $\sigma_c$ | H | V | D1 | D2 |
| 4 | 03.00±0.35 | NaN | NaN | NaN |
| 8 | 07.13±0.38 | NaN | NaN | NaN |
| 12 | 10.63±0.43 | 09.74±0.46 | NaN | NaN |
| 16 | 12.32±0.70 | 08.41±0.62 | 19.43±0.34 | 19.22±0.31 |
| 20 | NaN | 07.63±0.45 | 11.88±0.71 | 11.30±0.70 |
| 24 | NaN | 09.27±0.48 | 11.40±0.56 | 11.73±0.54 |
| 28 | NaN | 09.08±0.45 | 11.74±0.61 | 12.41±0.60 |

| Café Wall 7×11 (Mean Tilt ± StdErr) | | | | |
|---|---|---|---|---|
| $\sigma_c$ | H | V | D1 | D2 |
| 4 | 03.88±0.29 | NaN | NaN | NaN |
| 8 | 07.26±0.35 | 14.00±0.0 | NaN | NaN |
| 12 | 10.39±0.38 | 09.10±0.42 | NaN | NaN |
| 16 | 11.79±0.60 | 07.49±0.46 | 18.89±0.40 | 18.73±0.31 |
| 20 | NaN | 08.74±0.41 | 11.62±0.65 | 12.01±0.70 |
| 24 | NaN | 09.12±0.40 | 11.68±0.54 | 11.54±0.54 |
| 28 | NaN | 09.15±0.40 | 12.00±0.55 | 12.05±0.55 |

| Café Wall 11×7 (Mean Tilt ± StdErr) | | | | |
|---|---|---|---|---|
| $\sigma_c$ | H | V | D1 | D2 |
| 4 | 04.00±0.0 | NaN | NaN | NaN |
| 8 | 08.09±0.29 | NaN | NaN | NaN |
| 12 | 09.96±0.42 | 09.42±0.46 | NaN | NaN |
| 16 | 12.63±0.63 | 08.27±0.57 | 18.67±0.36 | 19.64±0.43 |
| 20 | NaN | 10.63±0.48 | 12.58±0.63 | 12.14±0.71 |
| 24 | NaN | 10.85±0.43 | 12.46±0.54 | 11.83±0.56 |
| 28 | NaN | 10.62±0.42 | 12.01±0.56 | 11.62±0.55 |

| Café Wall 11×11 (Mean Tilt ± StdErr) | | | | |
|---|---|---|---|---|
| $\sigma_c$ | H | V | D1 | D2 |
| 4 | 04.05±0.09 | NaN | NaN | NaN |
| 8 | 07.50±0.33 | 15.00±0.0 | NaN | NaN |
| 12 | 10.47±0.31 | 09.39±0.36 | NaN | NaN |
| 16 | 11.96±0.54 | 06.27±0.47 | 19.57±0.25 | 18.79±0.35 |
| 20 | NaN | 10.70±0.40 | 13.15±0.55 | 11.44±0.62 |
| 24 | NaN | 10.15±0.37 | 12.10±0.50 | 11.16±0.50 |
| 28 | NaN | 10.50±0.35 | 11.81±0.50 | 11.20±0.52 |

Fig 14 Left-column: Different configurations of the Café Wall pattern tested (Fig. 13) from Café Wall of 3×11 tiles on the Top to Café Wall of 11×11 tiles at the Bottom. Right-column: Mean tilts and the standard errors of tilt angles for each variation are summarised in the mean tilts tables for the four reference orientations of Horizontal (*H*), Vertical (*V*), and Diagonals (*D1*, *D2*) at seven different scales of the edge maps ($\sigma_c$ =4, 8, 12, 16, 20, 24, 28; Reproduced with permission from [23]).

The patterns explored here have the same size of tiles (200×200px) and mortar lines (8px) with these variations: Café Walls of 3×11, 5×5, 5×13, 7×11, 11×7, and 11×11 tiles, as shown in Fig 13. Looking at these variations, we see, for instance, a stronger tilt effect in the 5×13 tiles compared to the 5×5 tiles. Similarly, a stronger tilt effect is perceived in the variation of 7×11 tiles compared to a weaker tilts in the 11×7 tiles.

To eliminate the effect of *NumPeaks* on detected houghlines, in this experiment, we have selected *NumPeaks*=1000 to attain more accurate tilt measurements when the overall size of the Café Wall samples are not the same (similar to Section 3.2.1). We have also tested values above 1000 up to 5000 for this parameter, but we found empirically that there is no significant difference in the mean tilt results above *NumPeaks*=1000 for the samples tested and around four reference orientations. Increasing this value is computationally expensive and we need to keep a trade-off between the efficiency and the accuracy. The rest of the parameters are kept the same as to Figs 10 to 12 for the local and global investigations of tilt in variations of the Café Wall pattern. The Hough parameters are: *NumPeaks*=1000, *FillGap*=40, and *MinLength*=450 for all scales of the DoG edge maps ($\sigma_c$ = 4, 8, 12, 16, 20, 24, 28). Summary tables in Fig 14 present the quantitative mean tilts for the global tilt investigations on these configurations of the pattern (Fig 13). These include the mean tilts and the standard errors of the detected tilt angles around the four reference orientations (*H*, *V*, *D1*, *D2*).

The DoG outputs of these variations are the same across multiple scales, since the tiles and mortar lines have fixed sizes and the same set of parameters for the *Surround* and *Window ratios* (*s*, *h* respectively) have been used in the DoG model. Utilizing the Hough analytic pipeline for quantifying the tilt angles, we have measured slightly different tilts across the multiple scales of the edge maps of these variations around the four reference orientations. This is because Hough analyses more dominant lines (longest lines) first by applying the houghpeaks function prior to detecting lines with the houghlines function (MATLAB functions).

When the pattern is wider in horizontal direction such as the 3×11, 5×13 or 7×11 it seems that we see a stronger tilt effect along the mortar lines compared to the other variations. The quantitative mean tilts near horizontal orientation occur at fine to medium scales ($\sigma_c$ =4, 8, 12) of the edge maps. The near horizontal lines can be captured until scale 16 ($\sigma_c$ =16), with this mortar width (considering the same aspect ratio of the tile size to the mortar size), as well as the mid luminance of the mortar lines relative to the luminance of the tiles [21]). This can be seen also for the edge map of Café Wall of 3×8 tiles in Fig 2. There is a transient stage at scale 16 ($\sigma_c$ =16), connecting the detected near horizontal lines to the zigzag vertical line segments due to the arrangement of grouping of tiles in the zigzag vertical orientation at medium to coarse scales in the edge maps. The highest tilt range is shown in Fig 14 for the 5×13 configuration which is 3-10.6° at fine to medium scale ($\sigma_c$ =4, 8, 12), as expected. Then the variations of 7×11 (3.8-10.3°) and 11×11 (4-10.5°) come next, followed by the 3×11, 5×5 and 11×7 tiles. Considering the two square patterns (the 5×5 and 11×11 tiles), there are similar horizontal mean tilts, starting around 4.0° at the finest scale ($\sigma_c$ =4) and it is one degree wider in the 11×11 variation at scale 12 ($\sigma_c$ =12), ~10.5° compared to ~9.5° in the 5×5, but the differences are only significant for $\sigma_c$ =4.

The near vertical mean tilts at medium to coarse scales ($\sigma_c$ =20, 24, 28) show good results. The weakest vertical mean tilts correspond to the Café Wall of 3×11 tiles ranges from 7.1° to 7.5°. For the patterns of medium size height such as the 5×13 and 7×11 tiles, it is ~9-10°. It is in the highest range around 10.5° when the pattern is spread along the vertical orientation such as the 11×7 and 11×11 variations. This is nearly the same for the 5×5 tiles having the ratio of Height/Width=1, and with a maximum value for the 11×11 tiles (>10.6°). So the Café Wall of 5×13 tiles from the samples explored has the *strongest horizontal tilt* range while the 11×7 tiles shows the *strongest vertical tilt* range. It seems that there is a trade off in the mean tilts of the vertical and the horizontal, and for a stronger effect of vertical tilts, we encounter with a weaker horizontal tilts along the mortar lines. For the diagonal mean tilts the results show roughly similar deviations (in both positive and negative diagonals) at the coarse scales ($\sigma_c$ =24, 28) which is ~11-12° across the samples tested.

We note that the results reported here are based on our investigations on the number of lines detected at their angular positions and we have not considered any weights for the length of the lines in the mean tilts calculations. For the horizontal mean tilts this does not affect on the results, since at fine to medium scales the local tilted line segments (Twisted Cord elements) are extracted for all of these variations nearly the same with roughly similar size. The detected houghlines at scale 12 (medium-scale) for all variations tested have been provided in Fig 15, highlighting local tilts of the nearly horizontal Twisted Cords and small tilt deviations from the vertical. However, if a Café Wall pattern is more spread along the vertical orientation compared to the horizontal, then longer lines are detected with less deviation along the vertical at coarse scales (the whole tiles are present in the edge maps with no mortar cues left at these scales). Fig 16 clarify this more: the detected houghlines at scale 28 (the coarsest scale) have been presented for these variations, indicating the global tilts of the lines detected with zigzag vertical orientation. In fact, as we expect from the tilt estimation, deviations from the vertical increases as the lines found get shorter.

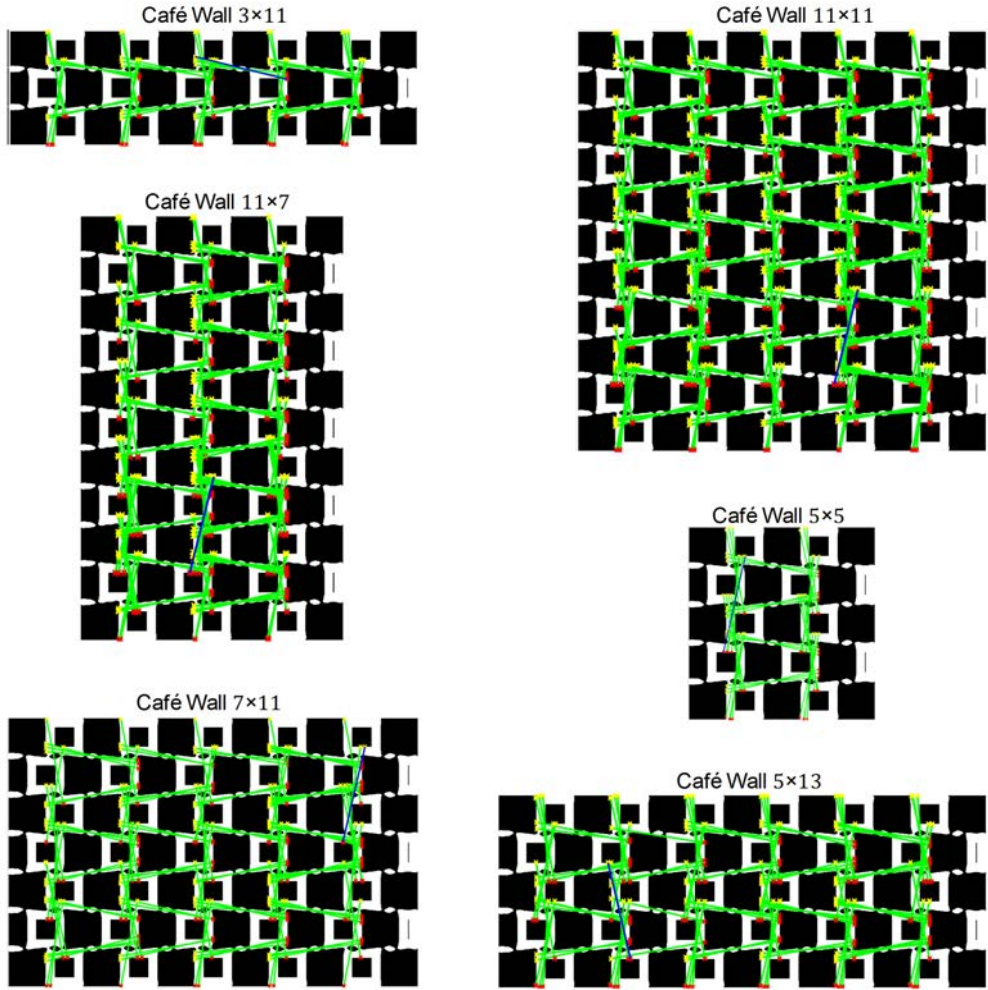

Fig 15. Detected houghlines displayed in Green, overlayed on the binary edge maps at scale 12 ($\sigma_c$ =12) of different configurations of the Café Wall pattern in Fig 13. Hough parameters are: *NumPeaks*=1000, *Threshold*=3, *FillGap*=40, and *MinLength*=450 (Reproduced with permission from [23]).

### 3.3 Comparison of local and global tilts in Café Wall illusion

We have shown in the last two sections that the mean tilt results with periphery appropriate parameterization are reliable and comparable with the previous results for foveal parameterization. The results for near horizontal global tilts in these variations are nearly the same as the local tilts detected in the foveal sample sets (Section 3.1.2) ~7° at scale 8. At scale 12, we have a higher tilt angle ~9.5-10.5° here compared to the local tilts around 8°. The results of the vertical and diagonal mean tilts are slightly larger than the predicted values for the foveal samples (7-10° for the vertical and 11-12° for the diagonal tilts here compared to ~6° for the vertical and 5-6° for the diagonal in the foveal samples). The results here seems more realistic in our perception of zigzag vertical lines at coarse scales considering the phase shift of rows of tiles in the Café Wall pattern (the deviations from the diagonal axes are more than 5°, considering the geometry of the pattern).

The quantitative modelling presented for the perceived tilt in the Café Wall illusion considering the foveal/local aspects as well as the peripheral/global view to the pattern leads us to achieve reliable results in our investigations. However, we illustrate some improvements to the current evaluation for future studies on the topic. First, for near horizontal mean tilts, although the tilt analysis pipeline in our model detects the local Twisted Cord elements as local tilt cues as shown in Fig 15 (for scale 12 for these variations), but it seems that in our perception of tilt, we intend to integrate these local tilt cues to construct a slanted continuous contour along the entire mortar as either diverging or converging [16, 28] tilt. Therefore, an edge integration technique is required for predicting a more precise value for the near horizontal tilts as we perceive tilts in the Café Wall. Second, in the investigated tilts around the vertical, we expect to see less deviations for the vertically spread configurations compared to the horizontally spread ones. However, the results showed the maximum vertical deviations for the Café Walls of 11×7 and 11×11 around 10.5° compared to 9° in others, except 7.5° for the Café Wall of 3×11. In the 3×11 tiles, we see more deviations around the diagonals compared to the rest of the configurations (as it is expected), which is ~12.5-14° compared to 11-12.5 °, also less deviations from the vertical due to the groupings of detected lines for the reference orientations. Our explanation of the results are getting clear by looking at the houghlines presented in Fig 16. In the Hough analysis, we have applied a same weight for all the detected lines. Therefore for the patterns that are spread along the vertical, although houghlines detects many longer lines with less deviations in the edge map, but houghpeaks lets more smaller line segments to be detected (up to the maximum value of *NumPeaks*), with more deviations from the vertical. For final validations of these results, psychophysical experiments are required as the priority of our future work. The results from psychophysical experiments lead us to assign weights to each scale and approximate tilt angles in our model based on the perceived tilt in real subjects.

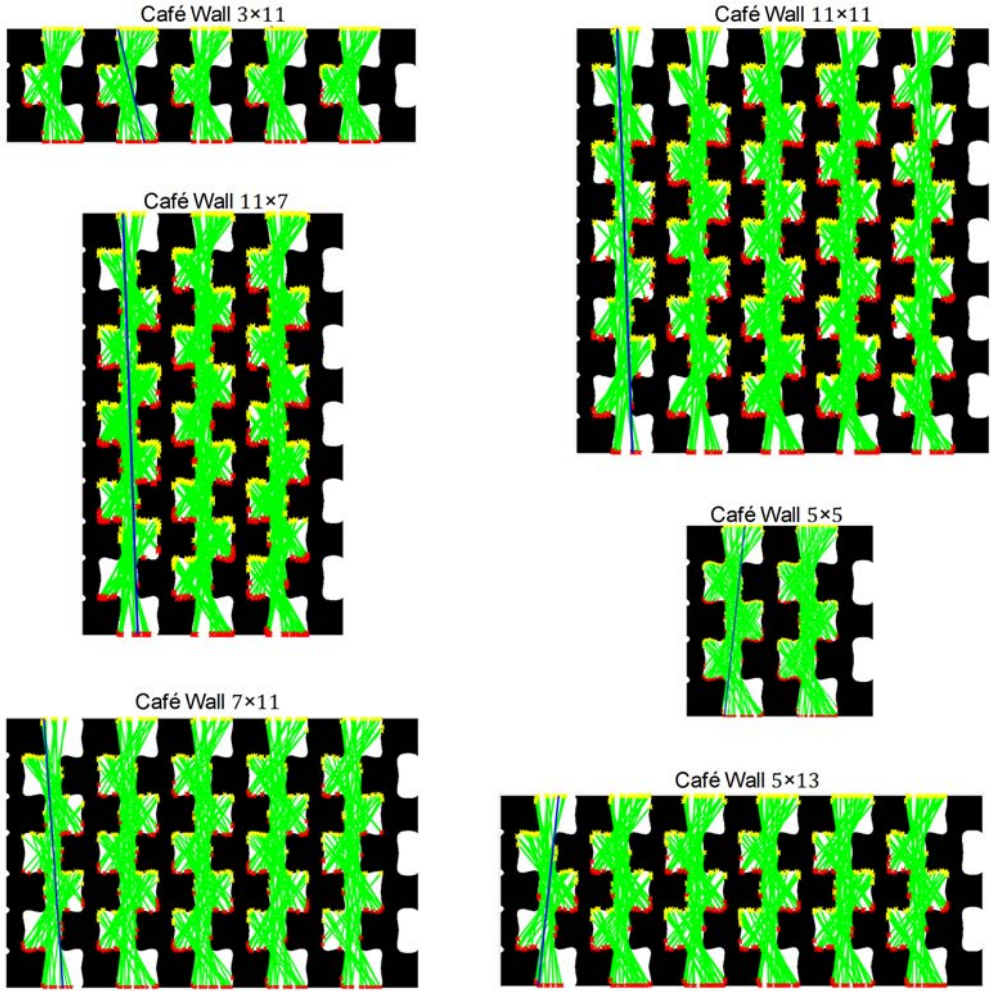

Fig 16 Detected houghlines displayed in Green, overlayed on the binary edge maps at scale 28 ($\sigma_c$ =28) of different configurations of the Café Wall pattern in Fig 13. Hough parameters are: *NumPeaks*=1000, *Threshold*=3, *FillGap*=40, and *MinLength*=450 (Reproduced with permission from [23]).

## 4 CONCLUSION

A low-level filtering approach [17, 19, 22, 23] have been explored here modelling the retinal/cortical simple cells in our early vision for revealing the tilt cues involved in the local and global perception of the Café Wall stimulus. The model has an embedded processing pipeline utilizing Hough transform to quantify the degrees of the inducing tilts appeared in the low-level representation for the stimulus in our model, referred to as the *edge map at multiple scales*.

The experiments reported have contributed new understanding of the relationship between the strength of tilt effect perceived in the Cafe Wall illusion as a function of eccentricity, that is whether a cell or edge is foveated or perceived in the periphery.

Different size/shape cropped samples of the Cafe Wall pattern were used to model the role of the shape and size of the fovea and larger samples tended to induce a larger number of longer shallower lines, particularly in the vertical dimension. When we foveate a particular cell we tend to see that as having more horizontal mortar boundaries, while those outside the fovea are perceived as having larger tilts. This is consistent with the larger tilts perceived at lower resolutions, modelling the periphery, and the almost horizontal tilts seen in the foveal region, corresponding to the centre of a larger pattern. This makes this a multiple scale model.

It is hypothesized that the multiple scale information from the retina is integrated later in the cortex into a true multiscale model, and that the Gestalt illusions result from the angle misperceptions that are already encoded in the retina. The quantitative predictions are based on the analysis of Hough transform of the edge maps here with promising results reported. This tilt investigation can be replaced by any more bioderived techniques, modelling mid- to high-level tilt integrations, capable of quantifying different degrees of tilt in variations of the gestalt view of the pattern, as we perceive the tilt differently in those variations.

We regard the publication of the predictions before running experiments to validate them as essential to the integrity of science. A priority in our research is psychophysical experiments to validate the predictions of the model.

## Acknowledgements

Nasim Nematzadeh was supported by an Australian Research Training Program (RTP) award for her PhD. The authors declare that there is no conflict of interest regarding the publication of this paper.